\documentclass[letterpaper]{article} 
\usepackage{aaai25}  
\usepackage{times}  
\usepackage{helvet}  
\usepackage{courier}  
\usepackage[hyphens]{url}  
\usepackage{graphicx} 
\usepackage{natbib}  
\usepackage{caption} 
\usepackage{algorithm}
\usepackage{algorithmic}
\usepackage{booktabs} 
\usepackage{amssymb}
\usepackage{multirow} 
\usepackage{amsmath} 
\usepackage{array}
\usepackage{subcaption}
\usepackage{xcolor}
\usepackage{colortbl}
\usepackage{newfloat}
\usepackage{listings}  
\definecolor{darkgreen}{RGB}{0, 100, 0} 

\DeclareCaptionStyle{ruled}{labelfont=normalfont,labelsep=colon,strut=off} 
\floatstyle{ruled}
\newfloat{listing}{tb}{lst}{}
\floatname{listing}{Listing}

\title{\textbf{FIRM}: Flexible Interactive Reflection ReMoval}
\author{
    Xiao Chen\textsuperscript{1,3}, Xudong Jiang\textsuperscript{2}, Yunkang Tao\textsuperscript{3}, \\
    Zhen Lei\textsuperscript{3,4,5}, Qing Li\textsuperscript{1}, Chenyang Lei\textsuperscript{3}\thanks{Corresponding authors: Chenyang Lei (leichenyang7@gmail\\.com), Zhaoxiang Zhang (zhaoxiang.zhang@ia.ac.cn)}, Zhaoxiang Zhang\textsuperscript{3,4,5}\footnotemark[1]
}
\affiliations{
    \textsuperscript{\rm 1}The Hong Kong Polytechnic University \textsuperscript{\rm 2}ETH Zurich\\
    \textsuperscript{\rm 3}Center for Artificial Intelligence and Robotics, HKISI-CAS\\
    \textsuperscript{\rm 4} Institute of Automation, Chinese Academy of Sciences\\
    \textsuperscript{\rm 5} University of Chinese Academy of Sciences\\
}

\usepackage{bibentry}

\begin{document}
 
\maketitle 
\begin{abstract}
\label{sec:abs}
Removing reflection from a single image is challenging due to the absence of general reflection priors. Although existing methods incorporate extensive user guidance for satisfactory performance, they often lack the flexibility to adapt user guidance in different modalities, and dense user interactions further limit their practicality.
To alleviate these problems, this paper presents {\textbf{FIRM}}, a novel framework for \textbf{F}lexible \textbf{I}nteractive image \textbf{R}eflection re\textbf{M}oval with various forms of guidance, where users can provide sparse visual guidance (e.g., points, boxes, or strokes) or text descriptions for better reflection removal. 
Firstly, we design a novel user guidance conversion module (UGC) to transform different forms of guidance into unified contrastive masks. The contrastive masks provide explicit cues for identifying reflection and transmission layers in blended images. Secondly, we devise a contrastive mask-guided reflection removal network that comprises a newly proposed contrastive guidance interaction block (CGIB). This block leverages a unique cross-attention mechanism that merges contrastive masks with image features, allowing for precise layer separation.
The proposed framework requires only 10\% of the guidance time needed by previous interactive methods, which makes a step-change in flexibility. 
Extensive results on public real-world reflection removal datasets validate that our method demonstrates state-of-the-art reflection removal performance. Code is avaliable at \url{https://github.com/ShawnChenn/FlexibleReflectionRemoval}.
\end{abstract}

\section{Introduction}
\label{sec:intro}

Image reflection removal refers to the task of eliminating unwanted reflections in images captured through glass. Specifically, the partially reflective glass superposes the scene of interest with reflections behind the observer, which reduces image contrast and potentially obscures important details.
Extensive research on image reflection removal primarily focuses on low-level and physics-based priors, such as gradient sparsity~\cite{levin2007user}, ghosting effect (where duplicate elements appear on thick glasses)~\cite{shih2015reflection_ghost}, and reflection blurriness~\cite{fan2017generic, Yang_2019_CVPR}. However, these methods often struggle with \textit{beyond-assumption} reflections (e.g., sharp reflections), due to the similarity in natural image statistics between transmission and reflection layers.

To alleviate the inherent ambiguity in layer separation, using auxiliary inputs as additional guidance has become a trend. Several works utilize multiple images or sensors to gather additional information about reflections, such as polarization images~\cite{eccv2018/Wieschollek,Lei_2020_CVPR,kong14pami,lyu2019reflection,Li_eccv20_PolarRR}, flash images~\cite{Lei_2021_RFC}, and multi-view images~\cite{DBLP:journals/tog/xue2015computational,dualviewrr,han2017reflection}. However, these methods require additional sensors or multiple captures, limiting their flexible applications in practice. 

\begin{figure*}[t]
  \centering 
  \vspace{-5pt}
  \includegraphics[width=0.85\linewidth]{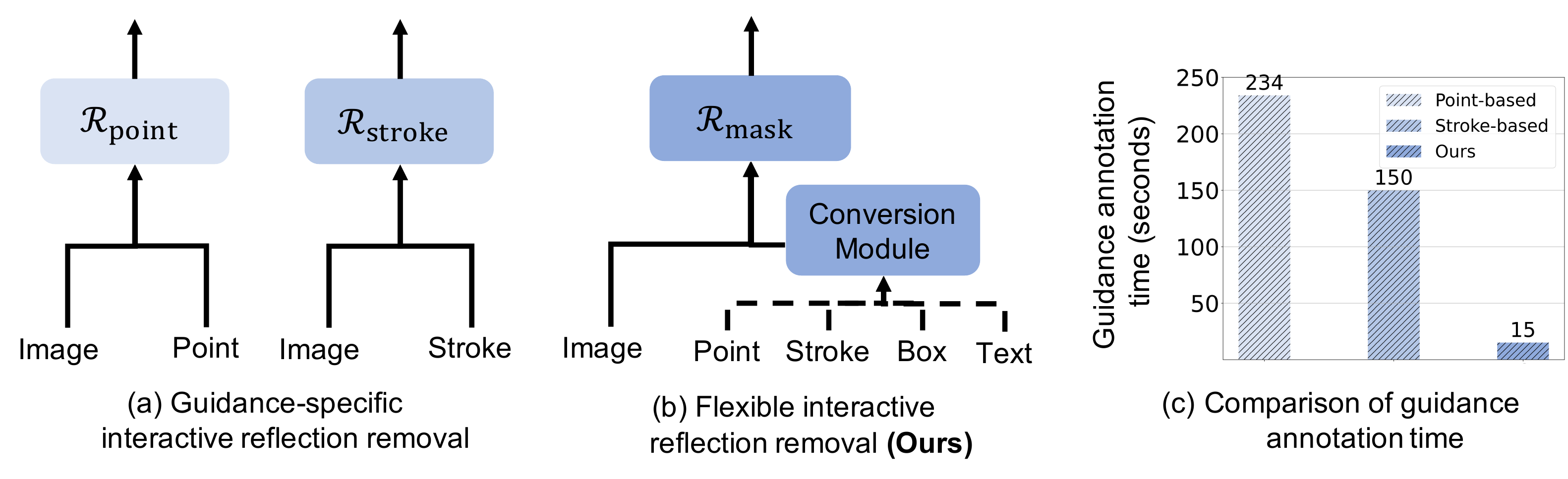} 
  \caption{\textbf{Comparison between previous interactive method~\cite{levin2007user, Zhang2020FastUS} and ours.} 
  \textbf{(a)} and \textbf{(b)} illustrate the structural differences. The previous methods are guidance-specific, with tailored reflection removal networks ($\mathcal{R_{\text{point}}}$, $\mathcal{R_{\text{stroke}}}$) for each guidance form(e.g., point or stroke). In contrast, our framework is flexible, utilizing a conversion module to accommodate various forms of guidance by transforming them into a unified ``segmentation mask". \textbf{(c)} Additionally, we compare the time cost of providing user guidance, where our method requires significantly less time per image than the results reported in previous works~\cite{Zhang2020FastUS}.}
  \vspace{-5pt}
  \label{fig:intro}
\end{figure*}

Interactive methods~\cite{levin2007user, Zhang2020FastUS} have also been studied, enabling reflection removal with more readily available human guidance, yet they exhibit significant limitations: i) support only a specific form of user guidance, and ii) require dense interactions for satisfactory performance, leading to high time costs. For instance, in~\cite{Zhang2020FastUS}, users draw dense strokes on the edge of reflection and background, resulting in nearly 150 seconds of time cost per image, as indicated in Figure~\ref{fig:intro}.

To address these limitations, we present FIRM, a novel interactive framework that supports flexible user guidance forms, including point, stroke, box, and text, for guiding reflection removal. As shown in Figure~\ref{fig:intro}, unlike previous interactive methods, our reflection removal network is not limited to specific guidance forms, as it incorporates a conversion module to unify various guidance into a mask format. Moreover, users can specify reflection and transmission layers with sparse guidance in an average of 15 seconds per image, significantly reducing the time cost from 234 seconds required by prior methods~\cite{levin2007user}.

Specifically, we propose a two-stage pipeline in FIRM. \textbf{Firstly}, we propose the \textbf{user guidance conversion} (UGC) module to convert different guidance into a unified format, that is, segmentation mask.
For text guidance, we adopt the text-based segmentation model~\cite{lai2023lisa}. For visual guidance (i.e., point, box, stroke), we develop a novel Segment Any Reflection Model (SARM) based on the Segment Anything Model (SAM)~\cite{kirillov2023SAM}, which freezes most parameters of SAM and updates only a learnable token and a feature selection block in the mask decoder. We do this because we observe that the original SAM falters in blended images when provided with sparse point prompts, as shown in Table~\ref{tab:sarm_metrics}. To address the performance degradation of SAM on blended images while maintaining its strong zero-shot capability, our SARM is trained using a lightweight parameter tuning strategy. Once trained, by prompting the UGC module with guidance on both transmission and reflection regions, we can obtain corresponding masks, which together form contrastive masks.
\textbf{Secondly}, we design a \textbf{contrastive mask-guided reflection removal network} that employs a novel contrastive guidance interaction block. This block enables the contrastive mask to interact with blended features and precisely separate out transmission and reflection features using cross-attention mechanisms.

To evaluate the efficacy of our proposed FIRM framework, we augment established benchmark datasets~\cite{zhang2018single,wan2017benchmarking} by incorporating additional user guidance, contributing to the first comprehensive interactive reflection removal dataset. Empirical results confirm that FIRM effectively improves reflection removal performance while requiring significantly less human guidance.
The main contributions of this work are threefold:

$\bullet$ We propose the first universal framework FIRM for interactive image reflection removal, supporting diverse flexible forms of guidance.  In particular, we develop the UGC module with a tailored segmentation model SARM, which enhances the ability to generate accurate reflection masks with sparse visual guidance. 

$\bullet$ We propose a novel reflection removal network that uses contrastive masks as additional guidance, employing a cross-attention mechanism to fuse transmission and reflection masks with blended image features. Extensive experiments demonstrate that it achieves superior reflection removal performance while requiring 10 $\times$ less time for annotating user guidance. 

$\bullet$ We contribute a comprehensive benchmark dataset for interactive image reflection removal, consisting of four forms of raw user guidance and their converted segmentation masks, facilitating further study in this field.
 
\section{Related Work}

\begin{figure*}[t]
    \centering
    \includegraphics[width=0.9\linewidth]{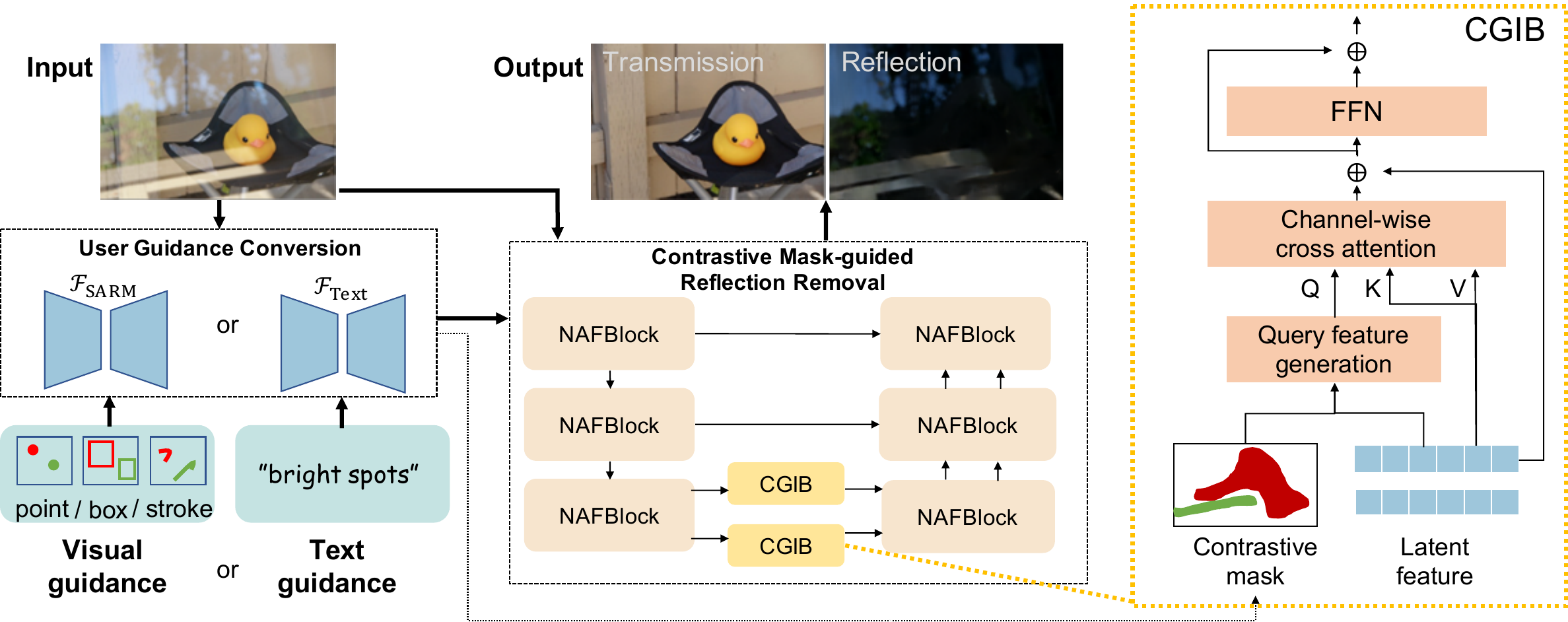}
    \caption{\textbf{Illustration of our proposed pipeline FIRM.} FIRM receives a blended image with diverse forms of user guidance, such as visual guidance or text descriptions. The user guidance conversion module (\textbf{UGC}) first transforms the raw input into contrastive masks with the user guidance. Then, the contrastive mask-guided network, incorporated with our designed Contrastive Guidance Interaction Block (\textbf{CGIB}) blocks, utilizes contrastive masks to separate the transmission and reflection layers from the blended input. (Detailed network configurations are provided in supplementary materials.) } 
   
    \label{fig:architecture}
\end{figure*}

\noindent\textbf{Single-image reflection removal.} Single-image reflection removal is challenging due to its ill-posed nature, which often leads to ambiguous decompositions, as explored in~\cite{wan2017benchmarking,wan2022benchmarking}. Traditional methods rely on defocused and ghosting cues. The defocus cue refers to reflections appearing blurry when focusing on the transmission layer due to depth disparity. Non-learning based methods ~\cite{Yang_2019_CVPR} exploit this by suppressing reflections with image gradient
statistics, while learning-based methods like~\cite{fan2017generic,zhang2018single} use these assumptions for data synthesis. The ghosting cue~\cite{shih2015reflection_ghost} is relevant for thick glass, identifies multiple reflections on the glass surface. However, these methods face limitations when these assumptions fail. Though several approaches employ GANs~\cite{Wen_2019_CVPR_Linear,Ma_2019_ICCV,DBLP:conf/nips/GoodfellowGAN14} or more accurate physical rendering methods~\cite{Kim_2020_CVPR} to mimic real
reflection distributions, or directly collect real-world data~\cite{zhang2018single,wei2019single_ERR,Li_2020_CVPR,lei2021categorized}, they still face challenges in covering diverse kinds of reflections~\cite{Lei_2020_CVPR,hu2023single,zhu2024revisiting}, underscoring the need for further research.

\noindent\textbf{Reflection removal with auxiliary inputs.} 
Alternative methods that use additional inputs have been explored. Motion-based techniques leverage multiple images to capture distinct motion characteristics, which aids in separating reflections. However, they require complex image capture setups and are limited by specific assumptions~\cite{guo2014robust,han2017reflection,li2013exploiting,liu2020learning,DBLP:conf/mm/Sun16RR,DBLP:journals/tog/xue2015computational,dualviewrr,chugunov2023neural}. Polarization-based methods leverage different polarization properties of reflection and transmission~\cite{Fraid1999,kong14pami,eccv2018/Wieschollek,lyu2019reflection, Li_2020_CVPR,Li_eccv20_PolarRR}. Flash/ambient image pairs have also been studied to handle reflections and shadows~\cite{agrawal2005removing_flash,chang2020siamese, lei2023robust}. These methods generally require additional equipment or specific image acquisition conditions. 
Interactive methods utilize user guidance, such as dense strokes or points, as additional input, but they require extensive user annotations~\cite{Zhang2020FastUS, levin2007user, chen2024towards}. In most recent work~\cite{zhong2024languageguided}, text descriptions are introduced as a form of high-level user guidance.
Our work diverges by offering a unified user guidance representation that accommodates various guidance forms, enhancing flexibility in user interactions. 

\section{Method}
\label{sec:method} 
\subsection{Overview}
In Figure~\ref{fig:intro}, we present the limited practicality of previous interactive methods, which motivates us to design two key objectives for the new framework: First, user guidance should be flexible and support various forms; Second, it ensures fast and convenient guidance annotation to further improve the interactive process. As illustrated in Figure~\ref{fig:architecture}, given the blended image 
and the raw user guidance in flexible modalities, the proposed FIRM framework predicts the underlying reflection and transmission images in two stages. First, the user guidance conversion module transforms the inputs into a unified contrastive mask, which captures prominent reflection and transmission region information.  
Then, the contrastive mask-guided transformer, built upon the Contrastive Guidance Interactive Block (CGIB) as its core component, integrates image features with the contrastive masks for precise decomposition.

\begin{figure*}[t]
    \centering
    \includegraphics[width=0.78\linewidth]{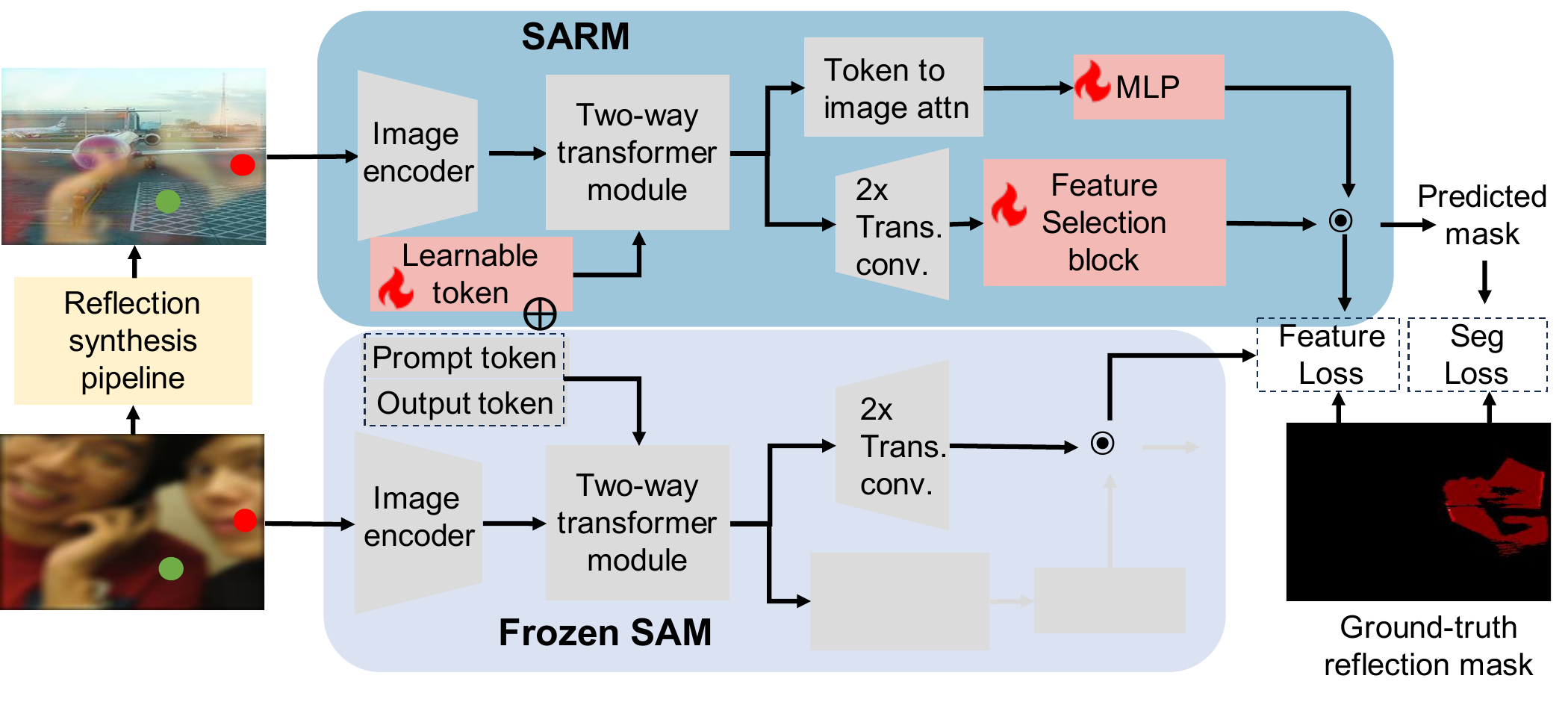}
    \caption{\textbf{Illustration of the training pipeline of SARM.} We introduce  learnable \textit{degradation-invariant token} and \textit{feature selection block} into the
    original SAM architecture, aiming for accurate mask prediction in blended images. To maintain the zero-shot capability of SAM~\cite{kirillov2023SAM}, only a limited number of parameters in the mask decoder are trainable, while the parameters of the image encoder and prompt encoder from the pre-trained SAM remain fixed.}
    \label{fig:sarm}
\end{figure*}

\subsection{UGC: User Guidance Conversion}
\label{sec:ugc}
To enhance the flexibility of utilizing various forms of user guidance, we introduce the UGC module to transform the user guidance $g^i \in \mathbf{G}$ into a unified mask format,
where $\mathbf{G}=\left\{g^i\right\}_{i=1}^{N}$ is the set of $N$ user guidance (i.e., point, box, stroke, text). The UGC module consists of interactive segmentation models, represented as $\mathcal{F}=\left\{\mathcal{F_{\text{SARM}}}, \mathcal{F_{\text{Text}}}\right\}$, where we propose a novel Segment Any Reflection Model (SARM) $\mathcal{F_{\text{SARM}}}$ for handling visual guidance and an off-the-shelf segmentation model $\mathcal{F_{\text{Text}}}$ for text guidance~\cite{lai2023lisa}.
The workflow can be formulated as:
\begin{align}
	\mathbf{M} = \mathcal{F}(\mathbb{S}(g^i), \mathbf{I}),
\end{align}
where $\mathbf{I}\in\mathbb{R}^{H\times W\times3}$ is the blended image and power set $\mathbb{S}(g^i)$ represents the corresponding mix set of user guidance, and segmentation mask $\mathbf{M} \in\{0,0.5,1\}^{H\times W\times1}$ contains distinct values for reflection (1), transmission (0.5) and non-annotated (0) areas.
Specifically, we propose SARM to address blended images while preserving the strong zero-shot learning capabilities of SAM~\cite{kirillov2023SAM}. 

\noindent\textbf{Preliminary of SAM.}
In the original SAM, the image encoder uses the Vision Transformer to process input images, and the prompt encoder handles sparse prompts (e.g., points, boxes) by converting them into suitable latent representations. The mask decoder then combines image and prompt embeddings with an output token using a two-way transformer module. It then applies transpose convolutions to upsample mask features and utilizes token-to-image attention to generate an output token for each mask. Finally, an MLP converts this output token into a dynamic classifier, which is multiplied with the mask features to produce the final segmentation mask.

\noindent\textbf{The tailored SARM.} 
To achieve more accurate segmentation on blended images using sparse visual prompts, we tailored SARM with minimal additional parameters to SAM, keeping SAM's image and prompt encoders fixed to maintain zero-shot capabilities while making two key modifications in the mask decoder.
First, we introduce the learnable degradation-invariant token.
This token
(size of \(1 \times 256\)) is concatenated with SAM's output tokens (size of \(4 \times 256\)) and prompt tokens (size of \(N_{\text{prompt}} \times 256\)) as the input to mask decoder.
Additionally, we incorporate a learnable three-layer MLP to generate dynamic weights, which are then used in a point-wise product with the mask features.
Second, we design the feature selection block to enhance prominent reflection features in blended images. This block takes intermediate mask features $\mathbf{F}\in \mathbb{R}^{c\times h\times w}$, first squeezing the spatial information into $\mathbf{F_{avg}} \in \mathbb{R}^{c\times 1\times1}$ via average pooling. It then employs a lightweight gating mechanism with sigmoid activation function $\sigma(.)$, as follows:
\begin{align}
\label{eq:maskfeature}
& \tilde{\mathbf{F}}=\sigma(\mathbf{W_1}(\operatorname{GELU}(\mathbf{W_0}(\mathbf{F_{avg}}))) \odot \mathbf{F},
\end{align}
where $\mathbf{W_0} \in \mathbb{R}^{c/r \times c}$, $\mathbf{W_1} \in \mathbb{R}^{c \times c/r}$ denote the learnable MLP weights. By exploiting the non-linear inter-channel relationship of mask features, the network learns to focus on channels that resemble salient features.

The training pipeline is shown in Figure~\ref{fig:sarm}. We first select one clear image and input it into SAM. This image is also used to synthesize the blended image, which is then fed into SARM. The proposed SARM is supervised by both feature-level and mask-level loss. For the mask-level segmentation loss, we
follow the configuration in~\cite{kirillov2023SAM}, combining Dice $\mathcal{L}_{\text{Dice}}$~\cite{sudre2017generalised} and Focal Loss $\mathcal{L}_{\text{Focal}}$~\cite{lin2017focal}. Additionally, we design a mask feature consistency loss to enhance the extraction of prominent reflection features. The overall loss function is as follows:
\begin{align} 
\label{eq:UGCLoss}
    \mathcal{L}_{\text{total}} = \mathcal{L}_{\text{Dice}}(\mathbf{M_{p}}, \mathbf{M_{gt}}) + \lambda_0 \mathcal{L}_{\text{Focal}}(\mathbf{M_{p}}, \mathbf{M_{gt}})\\
    \nonumber +\lambda_1 \mathcal{L}_{\text{MSE}}(\tilde{\mathbf{F}}_{p} \odot \mathbf{M_{gt}}, \tilde{\mathbf{F}_{gt}} \odot \mathbf{M_{gt}}),
\end{align}
where $\mathbf{M_{p}}$, $\mathbf{M_{gt}}$ denotes the predicted and ground-truth reflection mask, $\tilde{\mathbf{F_{p}}}$ and $\tilde{\mathbf{F_{gt}}}$ represent the mask features of SAM and SARM, $\lambda_0$ and $\lambda_1$ are hyper-parameters for different loss terms. 

\noindent \textbf{Constructing training data for SARM.} Since there is no public reflection segmentation dataset, we manually synthesize training data based on the COCO dataset~\cite{lin2014microsoft}. As illustrated in Algorithm~\ref{alg:synthesis}, we first apply the pipeline from~\cite{zhang2018single} to synthesize blended images using two clear images. Then we obtain a pseudo reflection instance mask $\mathbf{M_{r}}$ by traversing the instance mask with the highest values in the residual map, along with contrastive points located inside and outside the reflection mask. The proposed SARM is trained with blended images using contrastive points as prompts.

\begin{figure*}[t]
  \centering 
  \includegraphics[width=0.92\linewidth]{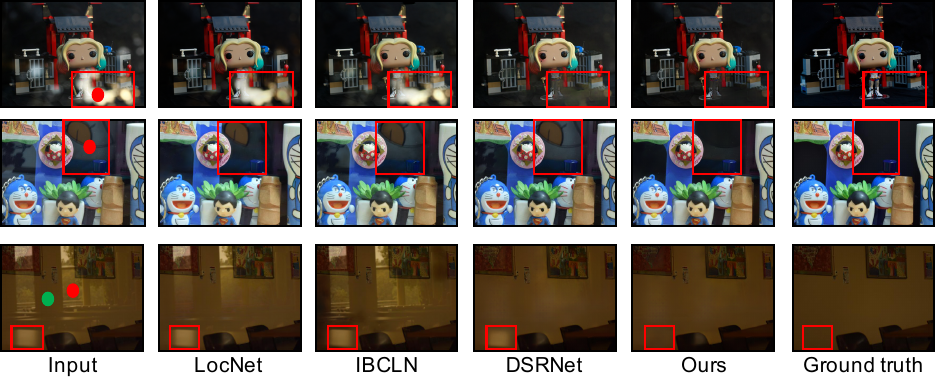} 
  \caption{\textbf{Qualitative comparison of estimated transmissions between representative single-image-based methods and ours on Real20 and SIR2 datasets.} Single-image-based methods struggle to remove sharp reflections. Our approach achieves much better reflection removal than baselines with very sparse point guidance on \textcolor{red}{reflection} and \textcolor{darkgreen}{transmisson} areas. }
  \label{fig:baseline_cmp}
\end{figure*}

\begin{algorithm}[t!]
\caption{Training data synthesis pipeline}
\label{alg:synthesis}
\textbf{Input}: Two clear RGB images $\mathbf{T}$, $\mathbf{R}$ and its instance mask $\mathbf{M}$ \\
\textbf{Output}: Blended image $\mathbf{I}$, reflection instance mask $\mathbf{M_{r}}$, contrastive points $\{p^{\text{pos}},p^{\text{neg}}\}$

\begin{algorithmic}[1]
\STATE $\mathbf{I} \gets$  Reflection\_Synthesis($\mathbf{T}$, $\mathbf{R}$) \\
\STATE $\mathbf{R^{'}} \gets \text{threshold}(\mathbf{I} - \mathbf{T}, 0)$ \# Residual map
\STATE max\_reflection\_value $\gets$ 0
\FOR{each instance $\mathbf{M_{i}}$ in $\mathbf{M}$}
    \STATE avg\_value $\gets$ MEAN($\mathbf{R^{'}} \cdot \mathbf{M_{i}}$)
    \IF{avg\_value $>$ max\_reflection\_value}
    \STATE max\_reflection\_value $\gets$ avg\_value
    \STATE  $\mathbf{M_{r}} \gets \mathbf{M_{i}} \cdot \mathbf{R^{'}}$
\ENDIF
\ENDFOR
\STATE Randomly sample a reflection point $p^{pos}$ from $\mathbf{M_{r}}$
\STATE Randomly select a transmission point $p^{neg}$ from the neighbour of $\mathbf{M_{r}}$
\RETURN $\mathbf{I}$, $\mathbf{M_{r}}$, $\{p^{\text{pos}}, p^{\text{neg}}\}$
\end{algorithmic}
\end{algorithm}

\noindent \textbf{Inference.} 
SARM supports points, boxes, or strokes as prompts. When the reflection area is labeled positively, we obtain the reflection mask; otherwise, we get the transmission mask. Stroke guidance is supported by uniformly sampling points along the stroke trajectory. Finally, we merge the reflection and transmission masks into a single mask, referred to as the contrastive mask.

\begin{table*}[t]
\begin{center}
{
\resizebox{0.98\linewidth}{!}{
\begin{tabular}{llcccccccccc}
\toprule         &        & \multicolumn{2}{c}{{ Real20 (20)}}       & \multicolumn{2}{c}{{ Object (200)}}     & \multicolumn{2}{c}{{Postcard (199)}}    & \multicolumn{2}{c}{{ Wild (55)}}             & \multicolumn{2}{c}{{ Average}}          \\ \cmidrule{3-12} 
\multirow{-2}{*}{Category}     & \multirow{-2}{*} {Method} & { PSNR}  & { SSIM}  & { PSNR}  & { SSIM}  & { PSNR}  & { SSIM}  & { PSNR}  & { SSIM}  & { PSNR}  & { SSIM}  \\ \midrule
 & Zhang et al.~\cite{zhang2018single}                 & { 22.55} & { 0.788} & { 22.68} & { 0.879} & { 16.81} & { 0.797} & { 21.52} & { 0.832} & { 20.08} & { 0.835} \\
 & BDN~\cite{Yang_2018_ECCV}                          & { 18.41} & { 0.726} & { 22.72} & { 0.856} & { 20.71} & { 0.859} & { 22.36} & { 0.830} & { 21.65} & { 0.849} \\
 & ERRNet~\cite{wei2019single_ERR}                       & { 22.89} & { 0.803} & { 24.87} & { 0.896} & { 22.04} & { 0.876} & { 24.25} & { 0.853} & { 23.53} & { 0.879} \\
 & IBCLN~\cite{Li_2020_CVPR} & { 21.86} & { 0.762} & { 24.87} & { 0.893} & { 23.39} & { 0.875} & { 24.71} & { 0.886} & { 24.10} & { 0.879} \\
 & RAGNet~\cite{li2023two}                      & { 22.95} & { 0.793} & { 26.15} & { 0.903} & { 23.67} & { 0.879} & { 25.53} & { 0.880} & { 24.90} & { 0.886} \\
 & DMGN~\cite{feng2021deep}                         & { 20.71} & { 0.770} & { 24.98} & { 0.899} & { 22.92} & { 0.877} & { 23.81} & { 0.835} & { 23.80} & { 0.877} \\
 &  Zheng et al.~\cite{zheng2021single}                 & { 20.17} & { 0.755} & { 25.20} & { 0.880} & { 23.26} & { 0.905} & { 25.39} & { 0.878} & { 24.19} & { 0.885} \\
 &  YTMT~\cite{hu2021trash}                         & { 23.26} & { 0.806} & { 24.87} & { 0.896} & { 22.91} & { 0.884} & { 25.48} & { 0.890} & { 24.05} & { 0.886} \\
 &  LocNet~\cite{dong2020location}                         & { 23.34} & { 0.812} & { 24.36} & { 0.898} & { 23.72} & { 0.903} & { 25.73} & { 0.902} & { 24.18} & { 0.893} \\
\multirow{-10}{*}{\begin{tabular}[c]{@{}l@{}}Single-image-based \\ methods\end{tabular}} & DSRNet~\cite{hu2023single}                         & { 23.85} & { 0.813} & { 26.38} & { 0.918} & { 24.56} & { 0.908} & { 24.79} & { 0.896} & { 25.46} & { 0.908} \\
\midrule
& $\text{Levin et al.}^{\dagger}$~\cite{levin2007user} &- & { -} &- &- &- &- &- &- &- & { 0.798} \\
& $\text{FGNet}$~\cite{Zhang2020FastUS} & { 24.21} & { 0.812} & { 24.97} & { 0.876} & { 22.84} & { 0.863} & { 25.23} & { 0.891} & { 24.07} & { 0.872}\\
 \multirow{-3}{*}{\begin{tabular}[c]{@{}l@{}}Interactive \\ methods\end{tabular}} & Zhong et al.~\cite{zhong2024languageguided} & { 24.05} & { 0.814} & { 26.41} & { 0.920} & { 24.62} & { 0.905} & { 26.20} & { 0.920} & { 25.53} & { 0.909}\\
\rowcolor{gray!30} & \textbf{Ours}  & { \textbf{26.88}}      & { \textbf{0.826}}      & { \textbf{26.80}}      & { \textbf{0.921}}      & \textbf{ 24.90}      & { \textbf{0.910}}      & { \textbf{29.71}}      & { \textbf{0.932}}      & { \textbf{26.34}}      & { \textbf{0.914}}      \\ 
\bottomrule
\end{tabular}}
\caption{\textbf{Quantitative comparison with baselines on Real20 and SIR2 dataset.} Our method trained with points (i.e., Ours-point) achieves the best performance on most evaluated datasets. We notice the reflection images in \textit{SIR2-Postcard} tend to be more blurry, which makes the performance difference smaller. In wild scenes like \textit{Real20}, \textit{SIR2-Object}, and \textit{SIR2-Wild}, where the reflections are sharper, the improvement of our approach is larger. For ~\cite{levin2007user}, we reference the results from ~\cite{Zhang2020FastUS}, indicated by $\text{}^{\dagger}$.}
\label{tab:baseline}
}
\end{center}
\end{table*}

\subsection{Contrastive Mask-Guided Reflection Removal}
\label{sec:Architecture2}
In the second stage, building upon a U-shaped encoder-decoder architecture, we propose a novel Contrastive Guidance Interaction Block (CGIB) that effectively incorporates guidance information from contrastive masks into the feature decomposition process. Our method differs from existing mask-guided methods~\cite{dong2021location} by integrating image features with auxiliary contrastive masks that provide more accurate region boundary information, enabling more precise layer separation.

Specifically, with the blended image $\mathbf{I} \in \mathbb{R}^{H \times W \times 3}$ and the converted contrastive mask $\mathbf{M} \in \mathbb{R}^{H \times W \times 1}$, we concatenate them along the channel dimension and feed it into the reflection removal network $\mathcal{R}$, which separates transmission layer $\hat{\mathbf{T}}$ and reflection layer $\hat{\mathbf{R}}$. The network primarily consists of NAFNet Blocks~\cite{chen2022simple} for feature extraction, incorporating CGIB into the middle layers for facilitating feature decoding.
The overall process is:
\begin{align}
   \{\hat{\mathbf{T}}, \hat{\mathbf{R}}\} = \mathcal{R}(\mathbf{I} \oplus \mathbf{M}, \mathbf{M}; \theta),
\end{align}
where $\theta$ denotes the network parameters of $\mathcal{R}$ and $\oplus$ denotes the concatenation operation. 

 The CGIB consists of three components, namely, Query Feature Generation, Channel-wise Cross Attention (CCA), and Feed-Forward Network (FFN). First, a query is generated using the contrastive mask. The contrastive mask is resized into $\mathbf{M^{'}} \in \mathbb{R}^{{h} \times {w} \times 1}$ to match the spatial dimensions of the blended features, followed by element-wise multiplication between the resized mask and blended features $\mathbf{I}_{\theta}$. 
 This step extracts the prominent reflection or transmission feature as queries, denoted as $\mathbf{Q}_{\theta}$. Note that, to handle the different resolutions of input images during inference, we resize $\mathbf{Q}_{\theta}$ to a fixed spatial dimension ${h'\times w'}$,
 Next, the CCA module uses the query feature $\mathbf{Q}_{\theta}$ as anchors to correlate similar components in the blended features $\mathbf{I}_{\theta}$. 
 The overall process of the CCA module is: 
\begin{align}
  \text{CCA}(\mathbf{Q}_{\theta}, \mathbf{K}, \mathbf{V}) = \mathbf{V}  \operatorname{Softmax}(\frac{\mathbf{Q_{\theta}} \mathbf{K}^{\top}}{\alpha}),
\end{align}
where $\mathbf{K}\in \mathbb{R}^{c \times {h'w'}}$ and $\mathbf{V}\in \mathbb{R}^{hw \times c}$ denote the blended image feature-generated key and value projections respectively, $\alpha$ is a temperature factor. The FFN component design strictly follows previous work~\cite{zamir2022restormer}. The whole network is trained with pixel-wise reconstruction loss in the image and gradient domain~\cite{hu2023single}, perceptual loss~\cite{wei2019single}, and exclusion loss~\cite{zhang2018single}. 

\section{Experiments}

\subsection{Dataset} Following the setting in~\cite{hu2023single}, the training data for reflection removal consists of 7,643 synthesized pairs from the PASCAL VOC dataset~\cite{everingham2010pascal} and 90 real pairs from~\cite{zhang2018single}. The proposed FIRM is trained using point guidance. For real data, we manually label one reflection and one transmission point per image. For synthetic data, we obtain contrastive points following the pipeline in Algorithm~\ref{alg:synthesis}. The test data includes \textbf{Real20} and \textbf{SIR2}~\cite{zhang2018single, wan2017benchmarking}. The \textit{SIR2} dataset~\cite{wan2017benchmarking} consists of three data splits: \textit{SIR2-Object}, \textit{SIR2-Postcard}, and \textit{SIR2-Wild}, each featuring distinct contents and depth scales.

\noindent\textbf{Flexible interactive reflection removal dataset.}
Since there is no publicly available evaluation dataset for interactive image reflection removal, we construct a comprehensive dataset that includes four forms of guidance. This dataset builds on the public reflection datasets Real20 and SIR2~\cite{zhang2018single, wan2017benchmarking}, where we further annotate prominent reflection and transmission areas in the blended images. 
We engage a team of annotators to label points, strokes, and bounding boxes on blended images. We then extract point coordinates from these annotations and feed them into the trained SARM as prompts to obtain corresponding segmentation masks. Text-guided segmentation masks are generated using the model~\cite{lai2023lisa}, with text descriptions manually labeled by the annotators.

\subsection{Implementation Details}
The proposed framework is implemented with PyTorch. During the training phase of SARM, only the proposed modules are optimized. Using point-based prompts, SARM is trained with a fixed learning rate of $0.0005$ for 50 epochs on 8 NVIDIA A100 GPUs. The batch size is set as 8. 
The reflection removal network is optimized using the Adam optimizer for a total of 200,000 iterations, with a batch size of 8 on a single A100 GPU. The initial learning rate is set to $10^{-3}$ and gradually reduce to $10^{-6}$ with the cosine annealing schedule ~\cite{loshchilov2016sgdr}.

\subsection{Evaluations on Reflection Removal}
We first compare the reflection removal performance of the proposed FIRM with two categories of methods, including single-image-based and interactive methods.
 
\noindent\textbf{Baselines.} \textbf{i)} \textit{Single-image-based methods}, including Zhang et al.~\cite{zhang2018single}, BDN~\cite{Yang_2018_ECCV}, ERRNet~\cite{wei2019single_ERR}, IBCLN~\cite{Li_2020_CVPR}, RAGNet~\cite{li2023two}, DMGN~\cite{feng2021deep}, Zheng et al.~\cite{zheng2021single}, YTMT~\cite{hu2021trash}, LocNet~\cite{dong2021location}, DSRNet~\cite{hu2023single}. \textbf{ii)} \textit{Interactive methods}~\cite{levin2007user, Zhang2020FastUS, zhong2024languageguided}. We retrain these methods on our training data for fair comparisons if their codes are available.

\noindent\textbf{Quantitative results.} In Table~\ref{tab:baseline}, we present the quantitative results of our approach and baselines on \textit{Real20}~\cite{zhang2018single} and \textit{SIR2}~\cite{wan2017benchmarking} dataset. We employ PSNR~\cite{huynh2008scope} and SSIM~\cite{wang2003multiscale} as metrics to evaluate the recovery quality of transmission layers.
Our proposed FIRM (with points as guidance), denoted as ``Ours,” consistently outperforms other methods in all data sets, showcasing its superior generalization ability and effectiveness. Notably, our method also surpasses the text-based interactive approach~\cite{zhong2024languageguided}. We speculate this arises from the recognizable layer ambiguity, where certain image layers lack corresponding language descriptions. In contrast, point annotations provide greater flexibility across diverse scenarios.
Additionally, our approach significantly reduces the reliance on dense point annotations, reducing the annotation number from 50~\cite{levin2007user} points per image to only 2 points. This sparse point guidance not only simplifies the user interaction process but also enhances the practicality of our method in real-world applications.

\begin{figure*}[t!]
  \centering 
  \includegraphics[width=0.9\linewidth]{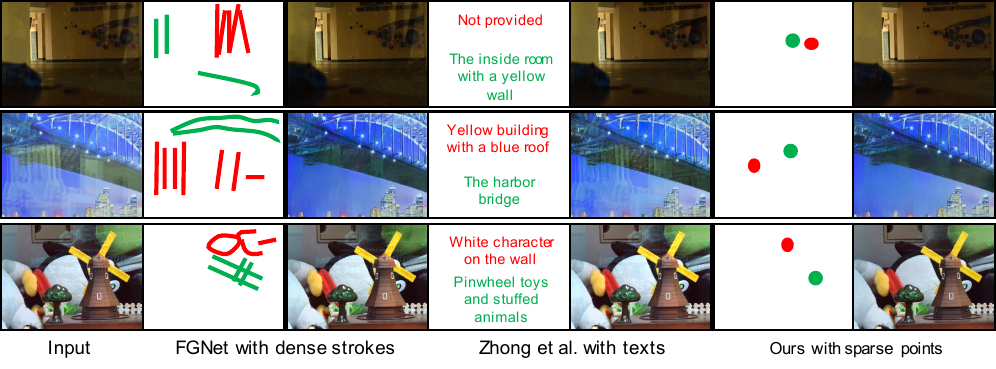} 
  \caption{\textbf{Qualitative comparison of predicted transmissions between state-of-the-arts interactive methods and ours on SIR2 datasets~\cite{wan2017benchmarking}.} The guidance for \textcolor{red}{reflection} and \textcolor{darkgreen}{transmission} regions is labeled with different colors. Our approach achieves superior reflection removal using just 2 sparse points.}
  \label{fig:interactive_cmp}
\end{figure*}

\noindent\textbf{Qualitative results.} We provide qualitative comparisons with single-image-based methods in Figure~\ref{fig:baseline_cmp}. As depicted, single-image methods often struggle to separate sharp reflections from the input image. 
For instance, bright spots on the cartoon dolls (row 1) and walls (row 3), and the pillow with orange patterns (row 2) have similar intensities as foregrounds. In contrast, our method is capable of producing high-quality transmission images.
We also present results from interactive methods in Figure~\ref{fig:interactive_cmp}, including FGNet~\cite{Zhang2020FastUS}, which requires dense scribbles and yields inferior results, and the method~\cite{zhong2024languageguided}, which uses text descriptions for reflections and transmission as additional guidance. For reflections lacking describable semantics (indicated as ``Not provided"), the text-based method struggles to identify them, while our approach uses just two or three sparse points to achieve significantly better visual quality.

\subsection{Ablation Study} 
\label{sec:ab}

We conduct ablation studies to validate the effectiveness of our proposed modules or designs, including \textbf{UGC}, \textbf{CGIB}, and the \textbf{Contrastive Mask}.  All model variants are trained from scratch using the same NAFNET-based architecture~\cite{chen2022simple} and evaluated on the Real20 and SIR2 datasets using points as additional guidance. These method variants include:
\noindent\textit{i) Blended Only}: Using only blended images as input to the network;
\noindent\textit{ii) Raw Point}: Without UGC and CGIB, raw points are directly combined with blended images;
\noindent\textit{iii) Raw Mask}: Without CGIB, the converted contrastive masks are directly combined with blended images as input;
\noindent \textit{iv) Reflection Mask}: Only the converted reflection mask is used for deep feature interaction in CGIB.
\noindent The average performance is shown in Table~\ref{tab:ab_firm}. Integrating raw points with the blended image directly or using blended image only yields inferior results, indicating the converted segmentation mask (UGC) is more effective for guiding removal. Directly combining converted masks with blended images also results in limited gains, emphasizing the importance of deep feature interaction (CGIB). 
Further, using reflection masks only for feature interaction cannot achieve optimal performance due to the lack of interaction cues.
\begin{table}[t!]
    \centering
    \resizebox{0.98\linewidth}{!}{ 
    \begin{tabular}{lcccccc}
        \toprule
        {Method} & NAFNet & UGC & CGIB &Contrastive Mask & PSNR &SSIM   \\ 
        \midrule
         Blended Only & \checkmark &  &  &  & 24.01 & 0.880 \\
         Raw Point  & \checkmark &  &  & & 24.05 & 0.886  \\ 
         Raw Mask & \checkmark & \checkmark & & & 25.70 & 0.907 \\ 
         Reflection Mask & \checkmark & \checkmark & \checkmark & & 26.07 & 0.902 \\ 
         \rowcolor{gray!30} \textbf{Ours} & \checkmark & \checkmark & \checkmark & \checkmark & \textbf{26.34} &\textbf{0.914} \\ 
        \bottomrule
    \end{tabular}
    }
    \caption{\textbf{Ablation study of the proposed FIRM.} Ablation results show that raw prompt-based methods underperform, while feature-level interactions with converted masks achieve better results across most datasets.}
    \label{tab:ab_firm}
\end{table}

We also evaluate the segmentation performance of the trained SARM on synthesized reflections using the COCO validation set~\cite{lin2014microsoft} and real-world reflections from SIR2~\cite{wan2017benchmarking}.
For comparison, we include RobustSAM~\cite{chen2024robustsam}, a recent model designed for degraded image segmentation. Unlike common degradations, reflections usually exhibit arbitrary patterns, which makes our proposed SARM more suitable for reflection segmentation, as shown in Table~\ref{tab:sarm_metrics}. 

\begin{table}[h]
    \centering
    \renewcommand{\arraystretch}{1.1} %
    \resizebox{0.98\linewidth}{!}{
    \begin{tabular}{lcccccccc}
        \toprule
        & \multicolumn{4}{c}{\textbf{Synthetic Data}} & \multicolumn{4}{c}{\textbf{SIR2}} \\
        \cmidrule{2-9}
        & \multicolumn{2}{c}{Reflection
        } & \multicolumn{2}{c}{Transmission} & \multicolumn{2}{c}{Reflection} & \multicolumn{2}{c}{Transmission} \\
        \cmidrule{2-9}
        \multirow{-3}{*}{Method} & IOU & Dice & IOU & Dice & IOU & Dice & IOU & Dice\\
        \hline
        Frozen-SAM & 0.306 & 0.469 & 0.840 & 0.905 & 0.326 & 0.432 & 0.803 & 0.879\\
        {RobustSAM} & 0.316 & 0.472 & 0.844 & 0.912 & 0.337 & 0.454 & 0.817 & 0.894\\
        SAM-decoder-ft & 0.369 & 0.505 & 0.854 &  0.923 & 0.449 & 0.574 & 0.824 & 0.903\\
        \rowcolor{gray!30}SARM & \textbf{0.429} & \textbf{0.565} & \textbf{0.857} & \textbf{0.926} & \textbf{0.541} & \textbf{0.702} & \textbf{0.829} & \textbf{0.906} \\
        \bottomrule
    \end{tabular}}    \caption{\textbf{Segmentation comparison on the synthetic data based on COCO validation set and real data on SIR2-dataset using point prompts.} ``-decoder-ft": finetuning the entire SAM mask decoder.}
    \label{tab:sarm_metrics}
\end{table}

\section{Conclusion}
This paper proposes a flexible interactive reflection removal approach that leverages human guidance in diverse forms as an auxiliary input. The user guidance conversion module, built upon a novel segment-any-reflection model, generates accurate reflection masks while preserving strong performance on clear images. Further, a Contrastive Guidance Interaction Block is designed in an encoder-decoder-based network to facilitate precise image layer separation using the generated masks, achieving superior reflection removal performance across various datasets. This highlights the significance of human guidance in addressing ambiguity in single-image reflection removal. Furthermore, we enhance existing public reflection removal datasets with sparse human annotations, facilitating further study.
\section{Acknowledgments}
This work was supported by the InnoHK program.
\clearpage
\bibliography{aaai25} 
\end{document}